\documentclass[runningheads]{llncs}
\usepackage[T1]{fontenc}
\usepackage[top=1.4in, bottom=1.4in, left=1.4in, right=1.4in]{geometry}
\usepackage{graphicx}
\usepackage{booktabs}
\usepackage[misc]{ifsym}
\newcommand{\corr}{(\Letter)}

\usepackage{multirow}
\usepackage{array}
\usepackage{amsmath,amssymb}
\usepackage{makecell}
\usepackage{subcaption}
\usepackage{xurl}
\usepackage{longtable}
\usepackage{tabularx}
\usepackage{ragged2e}
\usepackage[hidelinks]{hyperref}
\usepackage[utf8]{inputenc}

\newcommand{\meansd}[2]{\begin{tabular}[t]{@{}c@{}}#1\\[-0.2em]{\scriptsize (#2)}\end{tabular}}

\begin{document}

\title{Fairness Beyond Anonymization? Demographic Leakage in German LLM-Generated Résumés}

\titlerunning{Fairness Beyond Anonymization?}

\author{
Charlotte Leininger\inst{1}\thanks{These authors contributed equally to this work.} \and
Helena Veit\inst{1\ast} \and
Matthias A\ss{}enmacher\inst{1,2} \and
Andreas Bender\inst{1,2} \corr
}

\authorrunning{C. Leininger and H. Veit et al.}

\institute{
Department of Statistics, LMU Munich, Munich, Germany\\
\email{\{C.Leininger,Helena.Veit\}@campus.lmu.de}\\
\email{matthias@stat.uni-muenchen.de, andreas.bender@stat.uni-muenchen.de}
\and
Munich Center for Machine Learning (MCML), Munich, Germany
}

\maketitle              

\begin{abstract}
Large language models (LLMs) are increasingly integrated into AI-assisted hiring pipelines, including automated résumé generation and screening. Under the EU AI Act, the hiring domain is classified as high-risk, making fairness and transparency critical requirements. Existing work has primarily focused on explicit hiring decisions, while less attention has been paid to whether generated résumés themselves encode recoverable demographic information. In this work, we conduct a two-stage audit of demographic leakage in German-language LLM-generated résumés. First, we use ChatGPT (GPT-4o-mini), Gemini 2.5 Flash-Lite, and multiple scales of the open-weight Qwen 3 model family (4B, 8B, and 14B) to generate résumés from real anonymized job-matching profiles, systematically varying gender- and ethnicity-associated names while holding qualifications constant. Second, we simulate a downstream résumé screening scenario, where the generated résumés are first anonymized and gender-neutralized, before demographic leakage classifiers are trained on the resulting texts. We find that, despite these interventions, classifiers reliably distinguish between résumés generated with male and female names. This leakage is not driven by overtly gendered wording, but by subtle differences in the usage of semantically equivalent, formally gender-neutral terms in German. In contrast, ethnicity-related leakage remains comparatively weak across models. Our findings demonstrate that apparently neutral résumé generation can still preserve highly predictive demographic signals, raising concerns about anonymization-based fairness interventions in multilingual AI hiring pipelines.
\keywords{Fairness in AI \and Résumé Screening \and Multilingual NLP}
\end{abstract}

\section{Introduction}
\label{intro}
Recruitment and hiring processes such as résumé screening or candidate matching are increasingly supported by AI-assisted systems, including large language models (LLMs).
Estimates suggest that $99\%$ of Fortune 500 companies already rely on some form of automated support in hiring workflows \cite{hu201999}.
While such systems promise efficiency gains, they also raise concerns regarding fairness, discrimination, and the treatment of protected attributes such as gender and ethnicity.
In particular, the black-box nature of many AI-assisted hiring systems, combined with limited disclosure regarding the screening criteria and evaluation mechanisms used by these companies, creates substantial challenges for transparency and independent fairness auditing.
A prominent example involves an AI-based résumé screening system developed by Amazon, which was discontinued after it was found to disadvantage female applicants \cite{dastin2022amazon}.
Trained on historical hiring data reflecting male-dominated recruitment patterns, the system learned to penalize terms and experiences more frequently associated with women. 
Such examples illustrate how AI-assisted hiring systems may amplify or reproduce existing societal inequalities.
In response to these concerns, recent regulatory efforts, such as the EU AI Act\footnote{EU Artificial Intelligence Act \url{https://artificialintelligenceact.eu}}, classify employment-related AI systems as high-risk and require safeguards in the handling of sensitive and private information. 

While existing concerns about fairness in AI-assisted hiring primarily focus on explicit model decisions such as rankings or ratings \cite{fabris2025fairness,mujtaba2024fairness,koh_bad_2023,an_measuring_2025}, comparatively less attention has been paid to LLM-generated résumés themselves and their potential to introduce downstream screening and fairness risks.
As generative AI becomes increasingly integrated into recruitment workflows, LLMs are also used to summarize, refine, and structure candidate information in hiring and matching systems \cite{linkedin2024future,lo2025ai}.
At the same time, industry surveys suggest that a growing number of applicants rely on AI tools to write or improve résumés and cover letters \cite{resumebuilder2023}.
Consequently, LLM-generated résumé text may increasingly shape both applicant-facing and recruiter-facing stages of the hiring pipeline.
If LLM-generated résumés in this regard systematically encode demographic information through subtle linguistic patterns, downstream screening systems may pick up or even amplify these signals despite anonymization efforts. 

\begin{figure}[htbp]
    \centering
    \includegraphics[width=1\linewidth]{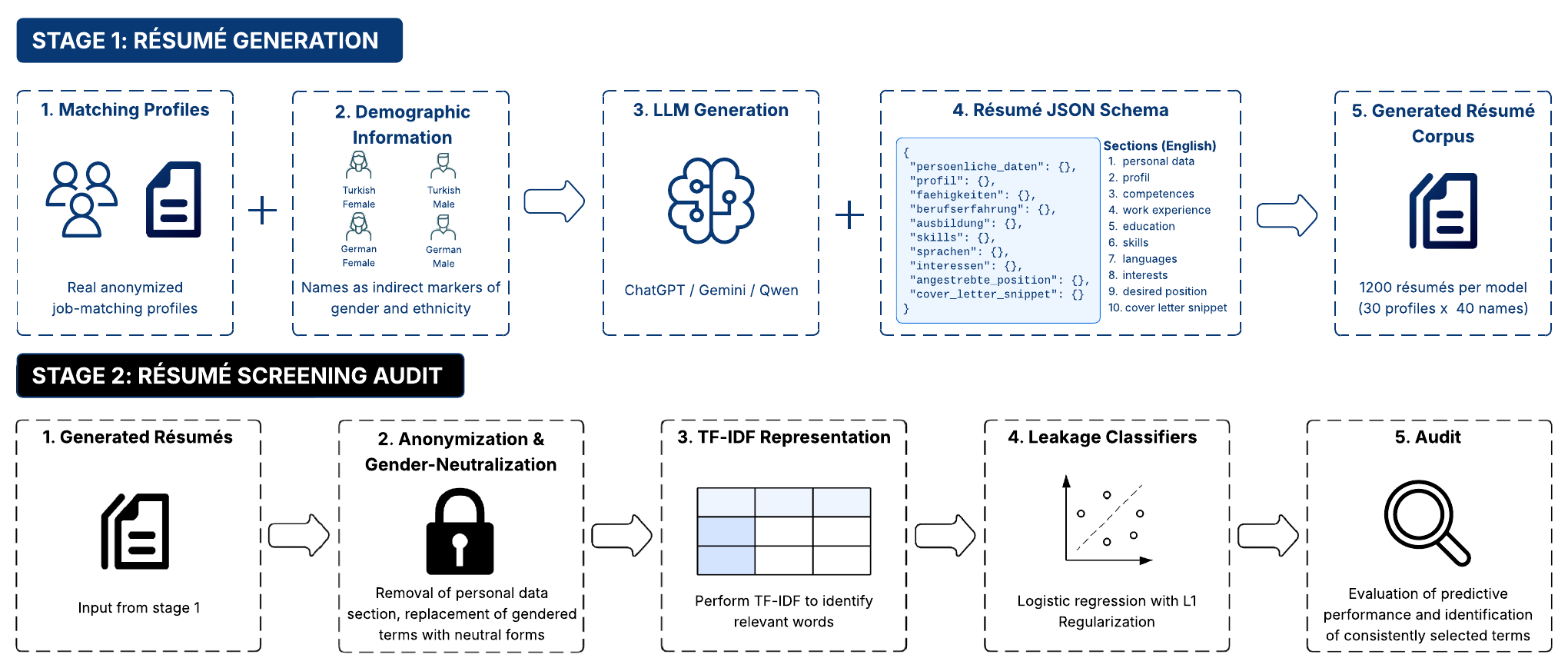}
     \caption[]{Overview of the two-stage audit pipeline. Stage 1 generates résumés from real anonymized job-matching profiles by systematically varying names as proxies for gender and ethnicity. Stage 2 assesses demographic leakage by anonymizing and gender-neutralizing the résumés before evaluating gender and ethnicity recoverability using TF-IDF representations and logistic regression classifiers.}
     \label{fig:audit_design}
 \end{figure}

\textbf{Contributions.} Against this background, we investigate whether LLM-gene\-rated German-language résumés contain latent demographic information that may create downstream fairness risks in automated screening systems. 
Using anonymized job-matching profiles provided by Chemistree GmbH\footnote{\url{https://www.chemistree.gmbh} is a German job-matching company that supports matching workflows by evaluating structured questionnaires and bringing together suitable matching partners. The present study was conducted as a student consulting project in cooperation with Chemistree GmbH.}, a German job-matching company, we conduct a controlled two-stage audit pipeline (for an overview, see Figure \ref{fig:audit_design}). 
In the first stage, we systematically vary gender- and ethnicity-associated names while holding all underlying qualifications constant and generate résumés using ChatGPT, Gemini, and multiple scales of the open-weight Qwen model family \cite{yang2025qwen3}. 
In the second stage, we assess downstream screening risks by evaluating whether demographic information can be recovered from the generated résumés using predictive models. 
Our findings demonstrate that demographic information, particularly gender, remains recoverable from generated résumés despite explicit anonymization and gender-neutralization procedures. 
Specifically, we identify linguistic signals associated with gender across all examined proprietary and open-weight LLM families. 
Our findings highlight a largely overlooked source of fairness risk in AI-assisted hiring, where in multilingual settings such as German, grammatical gender and language-specific lexical variation create additional opportunities for indirect demographic encoding. 
As a result, anonymization-based fairness interventions may not fully prevent demographic leakage in AI-assisted hiring pipelines.

\section{Related Work}
\label{related_work}

A rapidly expanding literature on AI-assisted hiring examines automated recruitment systems regarding fairness risks in applicant evaluation tasks \cite{fabris2025fairness,mujtaba2024fairness}. Existing studies include benchmark-based evaluations of candidate screening systems \cite{koh_bad_2023}, research on automated résumé evaluation \cite{an_measuring_2025}, as well as LLM-supported recruitment decisions \cite{gaebler_auditing_2024,armstrong_silicon_2024}, which all repeatedly identify demographic disparities related to gender or ethnicity. 

A common response to these fairness concerns in hiring involves interventions based on anonymization, which are designed to remove protected information prior to evaluation.
Beyond traditional blind recruitment approaches \cite{behaghel2015unintended}, recent work increasingly explores algorithmic anonymization and demographic removal strategies in AI-assisted hiring pipelines.
In particular, Parasurama and Sedoc \cite{parasurama-sedoc-2022-gendered} investigate résumé degendering and demographic removal techniques in English-language screening contexts.
Their work shows that linguistic preprocessing and anonymization may reduce demographic information available to downstream screening systems and therefore serve as potential fairness interventions in automated hiring.
Related work further explores fairness-aware hiring systems and synthetic recruitment settings intended to mitigate demographic bias during evaluation \cite{saldivar2025synthetic}.

At the same time, emerging work suggests that anonymization may not universally eliminate demographic information from recruitment-related text.
Behaghel et al. \cite{behaghel2015unintended} document unintended effects of anonymous recruitment procedures, while Tan et al. \cite{tan2026smallchangesbigimpact} show that subtle sociocultural markers in anonymized résumés can continue to produce demographic disparities in LLM-assisted hiring systems.
These findings raise broader questions regarding the conditions under which anonymization sufficiently prevents downstream demographic inference and bias.

However, beyond explicit demographic identifiers, studies show that language itself may also already encode socially meaningful demographic information \cite{wu2018gendered,qu_gender_2025}.
Gendered wording in recruitment materials has likewise been shown to distinguish between agentic language associated with stereotypically masculine traits and communal language associated with stereotypically feminine traits.
In particular, Gaucher et al. \cite{gaucher2011evidence} show that masculine- and feminine-associated wording in job advertisements may influence applicant behavior and reinforce labor-market inequalities.
Such findings suggest that demographic information may persist in text even when explicit identifiers are removed.

These concerns may be particularly relevant in multilingual and grammatically gendered languages. Most prior work on demographic language patterns and hiring fairness focuses on English-language settings, where grammatical gender plays a comparatively limited role.
In contrast, German contains gendered morphology and language-specific lexical variation that may create additional pathways for indirect demographic encoding \cite{motschenbacher2014grammatical}.
Our work addresses this gap by auditing anonymization-based screening interventions for German-language LLM-generated résumés and examining the recoverability of demographic gender and ethnicity signals across models after deliberate anonymization.

\section{Résumé Generation}

\label{sec:resume_generation}
The first stage of our two-stage audit pipeline consists of generating résumés from anonymized job-matching profiles using multiple LLMs (see Stage $1$ of Figure \ref{fig:audit_design}).
To isolate demographic effects, we hold underlying profile information constant while gender- and ethnicity-associated signals are varied exclusively through applicant names.
This design enables systematic comparisons of generated résumé content under otherwise identical conditions and forms the basis for the downstream demographic leakage audit.
All code, regarding both stages of the audit, is available on GitHub\footnote{\url{https://github.com/helenaveit/Fairness-Beyond-Anonymization}}.

\subsection{Matching Profile Data}
\label{sec:data}

The résumé generation is based on anonymized applicant information obtained from structured job-matching profiles provided by the German matching platform Chemistree.
Using real profiles introduces natural variation while reflecting practical LLM-supported recruitment workflows.
The profiles contain no personal identifiers such as names, age, gender, or ethnicity.
This ensures that any demographic signal present in the generated résumés originates solely from the assigned names.
Given the sensitivity of job-matching data, details on data sourcing, consent, and anonymization are provided in the \nameref{sec:ethical_statement} at the end of the paper.
The full dataset comprises 5,932 profiles, of which 3,542 are fully completed.
From this subset, we draw a random sample of $N_p = 30$ profiles for the experimental setup.

\paragraph{Questionnaire Content.}

The underlying questionnaire is designed for job matching and captures applicant characteristics across three broad domains: hard skills, soft skills, and applicant preferences. Details regarding the questionnaire are provided in Appendix~A.
Although the profiles differ from conventional résumés, this does not affect the present audit, which focuses on demographic leakage rather than reconstructing complete career histories.

\subsection{Name Data Set}
\label{sec:name_data}
To introduce controlled demographic variation into the résumé generation process, we construct an external set of $N_n=40$ full names and combine these with the sampled matching profiles.
The full list of names can be found in the Appendix B.

\paragraph{Ethnicity and Gender Dimensions.}
We operationalize gender as binary, as it is inferred from names rather than self-reported.
Accordingly, each name $i=1,...,N_n$ is associated with a gender attribute $g_i\in\{\text{male},\text{female}\}$.
We additionally investigate ethnicity as a second potential source of bias, associating each name with an ethnicity attribute $e_i\in\{\text{German}, \text{Turkish}\}$.
We focus on Turkish-associated names since individuals with Turkish migration backgrounds constitute one of the largest migrant-origin populations in Germany and prior labor-market studies document persistent discrimination against Turkish-associated applicants in hiring contexts \cite{Kaas2012Ethni-21452,kohler2025field}.   

\paragraph{Sampling.}
To mitigate name-specific bias, we sample ten names per demographic subgroup rather than relying on a single representative identity.
Since the original matching profiles are anonymized and contain no age information, all applicants are assigned an age of 40. First names are drawn from frequency statistics for birth year 1985 to align with the assigned age, using common German and Turkish names separated by gender.\footnote{Behind the Name: Baby Name Statistics: \url{https://www.behindthename.com/names}}
Surnames are drawn from German and Turkish frequency statistics and paired across genders such that each male--female pair shares a surname.\footnote{Nachnamen.net: Surname Frequency Statistics: \url{https://www.nachnamen.net}}

\subsection{Pipeline Workflow}
\label{sec:pipeline}
    
We employ an automated generation pipeline to systematically construct the résumé corpus for all demographic conditions and LLMs.
Each of the $N_p = 30$ sampled matching profiles is paired with each of the $N_n = 40$ sampled names, yielding $N_{p\times n}=1{,}200$ unique profile--name combinations per model.

For each combination, the LLM receives the anonymized matching profile together with a single assigned name and is instructed to generate a realistic German-language résumé.
All profile information remains fixed across demographic conditions, while gender- and ethnicity-associated signals vary exclusively through the assigned name.
This controlled counterfactual design isolates demographic variation from applicant qualifications.
The resulting résumé corpus forms the final basis for the downstream anonymization and demographic leakage audit.

\subsection{Technical Implementation}

Experiments are conducted using GPT-4o-mini and Gemini 2.5 Flash-Lite, alongside three parameter scales of Qwen 3: \texttt{Qwen3-4B}, \texttt{Qwen3-8B}, and \texttt{Qwen3-14B} \cite{yang2025qwen3}.
Including both proprietary and open-weight models enables assessment of whether demographic leakage patterns generalize across model families and scales.
    Temperature is fixed at 1 for all models to preserve natural generation variability while maintaining coherent résumé outputs.
    While for ChatGPT \texttt{top\_k} cannot be manually set, Gemini has a fixed \texttt{top\_k=64}.
    For all Qwen 3 variants, we use both the suggested \texttt{top\_k=20} from their best practices, as well as \texttt{top\_k=64} to allow direct comparison with Gemini's fixed setting while additionally assessing robustness to decoding choices.

\paragraph{Prompt Structure and Output Formatting.}
The full prompt is provided in Appendix~C. It is separated into a \texttt{SYSTEM} and \texttt{USER} part.
The \texttt{SYSTEM} prompt defines the task, output language, stylistic conventions, and required JSON schema, and explicitly instructs the model to make realistic assumptions when profile details are missing rather than inserting placeholders.
The \texttt{USER} prompt supplies the matching profile and name for each call, along with a short example of the expected output structure.

\paragraph{Résumé Schema Design.}
Generated résumés follow a standardized schema broadly reflecting conventional résumé structure, including the sections personal data, a short profile description, competences, work experience, education, skills, languages and interests. 
In addition, the schema further includes a section for the desired position and a short cover letter snippet to encourage more free-form text generation.

\section{Résumé Screening Audit}
\label{sec:resume_screening}

In the second stage of the pipeline, we assess whether demographic information remains recoverable from LLM-generated résumés after anonymization and linguistic preprocessing (see Stage $2$ of Figure \ref{fig:audit_design}).
Rather than modeling a complete hiring process, this stage simulates an anonymized downstream screening setting and evaluates demographic prediction as an audit measure of linguistic leakage.
This setup reflects existing fairness- and privacy-oriented screening practices, where anonymization is commonly used to reduce the influence of protected attributes and is aligned with broader regulatory objectives of the EU AI Act for high-risk employment systems.
If gender or ethnicity remains predictable from processed résumé text despite anonymization and gender-neutralization, this indicates that demographic information continues to be indirectly encoded in the generated résumés.
To this end, we first apply anonymization-based and linguistic normalization procedures and subsequently train demographic leakage classifiers based on TF--IDF representations of the résumé text.

\subsection{Anonymization}
\label{sec:screening_scenario}

\paragraph{Removal of Personal Information.}

We remove the top-level JSON field regarding the personal data section, which contains identity-related information such as names, contact details, email addresses, and LinkedIn profiles.
These fields are directly associated with demographic attributes and would constitute trivial sources of gender and ethnicity information.
Additionally, names are removed from all remaining résumé sections whenever mentioned.

\paragraph{Gender-Neutralization of German Word Forms.}
     We additionally neutralize gender specific lexical forms characteristic of the German language.
     German occupational and role nouns often encode grammatical gender through suffixes or gender-marked lexical alternatives, potentially providing trivial cues for demographic prediction.
     To prevent leakage classifiers from relying on such overt signals, we apply a rule-based gender-neutralization procedure (Table~\ref{tab:gender_stems}).
     For nouns with feminine suffixes such as \textit{-in}, we map feminine forms to their neutral or canonical counterpart (e.g., \textit{Beraterin} $\rightarrow$ \textit{Berater}, \textit{eng.} both \textit{consultant}).
     For gender-marked lexical pairs such as \textit{Kaufmann}/\textit{Kauffrau} (\textit{eng.} counterpart \textit{salesman}/\textit{saleswoman}) or \textit{Fachmann}/\textit{Fachfrau} (\textit{eng.} both \textit{specialist}), we transform both forms into a shared gender-neutral representation (e.g., \textit{Kauf-mann/-frau}, \textit{Fach-mann/-frau}).
     This preprocessing step removes explicit grammatical gender markers, ensuring that any remaining demographic recoverability reflects more subtle linguistic patterns rather than overt gendered wording alone.

\begin{table}[htbp]
\centering
\caption{Examples of gender-stem normalization.}
\label{tab:gender_stems}

\setlength{\tabcolsep}{14pt}

\begin{tabular}{lcl}
\toprule
Original word &  & Gender-neutral form \\
\midrule
Beraterin (\emph{female consultant}) 
& $\rightarrow$ & 
Berater (\emph{consultant}) \\

Kaufmann (\emph{salesman}) 
& $\rightarrow$ & 
Kauf-mann/-frau (\emph{salesperson}) \\

Kauffrau (\emph{saleswoman}) 
& $\rightarrow$ & 
Kauf-mann/-frau (\emph{salesperson}) \\

Fachmann (\emph{male specialist}) 
& $\rightarrow$ & 
Fach-mann/-frau (\emph{specialist}) \\

Fachfrau (\emph{female specialist}) 
& $\rightarrow$ & 
Fach-mann/-frau (\emph{specialist}) \\
\bottomrule
\end{tabular}
\end{table}

\paragraph{Text Preprocessing.}
Following anonymization and gender-neutralization, we apply standard text preprocessing procedures to the résumé texts in Python.
German stop words are removed using the \texttt{stop\_words} package, and lemmatization is performed using spaCy's German language model \texttt{de\_core\_news\_md} \cite{honnibal2020spacy}, mapping inflected forms to their canonical base forms.
We restrict the analysis to unigram features and discard rare terms by retaining only vocabulary items appearing in at least ten résumés.
These preprocessing steps reduce lexical sparsity and ensure that demographic leakage is evaluated on normalized linguistic representations rather than idiosyncratic word forms.

\subsection{Demographic Leakage Classifiers}
\label{sec:leakage_classifiers}

\paragraph{Classification Setup.}

Following prior work, we treat demographic prediction performance as a diagnostic measure of demographic leakage rather than as a predictive objective (see, e.g., \cite{parasurama-sedoc-2022-gendered,qu_gender_2025,wu2018gendered}).
High predictive performance indicates that demographic information remains encoded in the generated résumé texts despite anonymization and linguistic preprocessing.
We evaluate classifier performance using accuracy (ACC) and area under the ROC curve (AUC). Under this balanced leakage audit framework, values of $\text{ACC}=0.50$ and $\text{AUC}=0.50$ correspond to a random-guessing classifier and thus indicate an absence of detectable demographic leakage.

\paragraph{Term Frequency--Inverse Document Frequency.}

We represent résumé texts using term frequency--inverse document frequency (TF--IDF) features.
Defining the corpus at the résumé level yields a collection of $N_d = 1{,}200$ documents $d$ per model.
The TF--IDF of a term $t$ in a document $d$ is defined as the term frequency $\text{tf}_{t,d}$ weighted by the inverse document frequency $\text{idf}_t$ of the term across the corpus.

\begin{equation}
\begin{aligned}
\text{tfidf}_{t,d}
&=
\text{tf}_{t,d}
\cdot
\text{idf}_{t}
\
&=
\text{tf}_{t,d}
\cdot
\left(
\log
\left(
\frac{N_d}{\text{df}_{t}}
\right)
+1
\right)
\end{aligned}
\end{equation}

Here, $\text{df}_{t}$ denotes the number of résumés containing term $t$, and $N_d$ the total number of résumés in the corpus. We compute TF--IDF representations using the \texttt{TfidfVectorizer} implementation from \texttt{sklearn.feature\_extraction.text} in Python \cite{scikit_learn}.
Each résumé is thereby represented as a TF--IDF vector, forming the basis for the demographic leakage classifiers.
We intentionally employ TF--IDF representations to preserve interpretability at the lexical level.
While more complex embedding-based representations may capture additional forms of demographic information, TF--IDF enables direct identification of the specific linguistic terms contributing to demographic recoverability, which aligns with the primary objective of the present audit.

\paragraph{Logistic Regression with L1 Regularization.}

We employ logistic regression with $L_1$ regularization \cite{hastie2009elements,tibshirani1996regression} as the classification model, which is given by
\begin{equation}
P(g_d \mid \mathbf{x}_d;\mathbf{\theta}) = \sigma\left(\mathbf{\theta}^\top \mathbf{x}_d\right), \quad \mathbf{\theta}^* = \arg\min_{\mathbf{\theta}} \sum_{d=1}^{N_d} \mathcal{L}_{CE}(g_d, \hat{g}_d) + \lambda \|\mathbf{\theta}_{1:}\|_1.
\end{equation}
Here, $g_d \in \{\text{male}, \text{female}\}$ denotes the gender attribute of the name assigned to résumé $d$, with ``$g_d = \text{male}$'' as the positive class. $\hat{g}_d = P(g_d \mid \mathbf{x}_d; \boldsymbol{\theta}) \in [0,1]$ denotes the predicted probability, $\mathbf{x}_d$ the TF--IDF vector of résumé $d$, $\mathcal{L}_{\text{CE}}$ binary cross-entropy loss, $\sigma$ the sigmoid function, $\lambda$ the regularization parameter controlling solution sparsity, and $\|\boldsymbol{\theta}_{1:}\|_1$ the $L_1$ penalty applied to all parameters excluding the intercept $\theta_0$. The analogous formulation applies for ethnicity prediction, replacing $g_d \in \{\text{male}, \text{female}\}$ with $e_d \in \{\text{German}, \text{Turkish}\}$, where ``$e_d = \text{German}$'' is the positive class. Note that $g_d$ and $e_d$ are résumé-level restatements of the name-level attributes $g_i$ and $e_i$ defined in Section~\ref{sec:name_data}, inherited through the profile--name assignment.

For each task and model, we perform an 80/20 train--test split using group-wise partitioning with respect to the profile number.
This ensures that résumés derived from the same underlying matching profile do not appear in both training and test sets, preventing the risk of learning profile-specific characteristics rather than systematic demographic patterns.
The regularization parameter $\lambda$ is selected via grouped 3-fold cross-validation on the training data, again partitioned by profile number.
We employ $L_1$-regularized logistic regression primarily as a sparse feature selection method, allowing identification of linguistic terms most strongly associated with gender and ethnicity while shrinking less informative coefficients toward zero.
Model selection follows the 1-SE rule, selecting the simplest model whose mean cross-validated deviance lies within one standard deviation of the minimum.
Because the number of underlying profiles is limited ($N_p = 30$), we employ a 30 times repeated hold-out split (80/20) in order to identify linguistic signals that are selected consistently across runs and are therefore less likely to reflect split-specific artifacts.
All models are fitted using the \texttt{glmnet} package \cite{friedman2010regularization} in \texttt{R}, with features standardized prior to model fitting.

\section{Results}

We first verify that the generated résumé corpus is suitable for the downstream audit before turning to the central leakage analysis.
The main results then assess whether gender- and ethnicity-related information remains recoverable after anonymization and linguistic preprocessing, and which linguistic signals are selected by the leakage classifiers (Section \ref{res:screening}).
Finally, we examine whether these patterns change across Qwen 3 model sizes (Section \ref{res:qwen_scaling}).

\paragraph{Generated LLM-outputs.}
Generated résumés required minor postprocessing to correct model-specific JSON formatting inconsistencies. These issues were most common for Qwen models and Gemini, whereas ChatGPT produced no parsing errors. After normalization, all generated résumés were suitable for downstream analysis.
Regarding résumé length distributions across models, Gemini generated the longest résumés on average, followed by the Qwen models, while ChatGPT produced substantially shorter outputs. No notable differences were observed across gender or ethnicity groups. Per-gender and per-section length statistics are reported in Appendix E.

 \begin{figure}[t]
     \centering
     \includegraphics[width=0.48\textwidth]{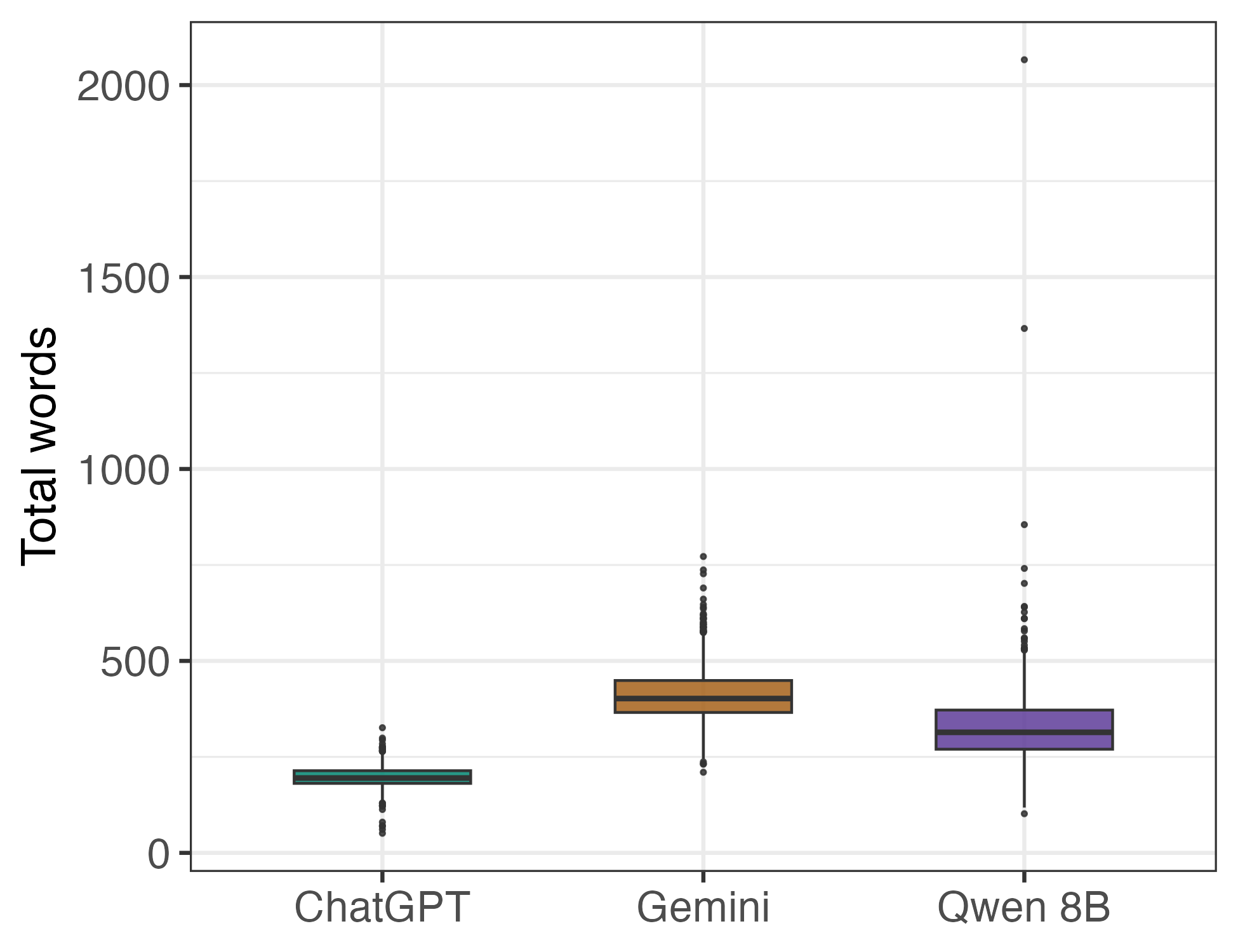}
     \hfill
     \includegraphics[width=0.48\textwidth]{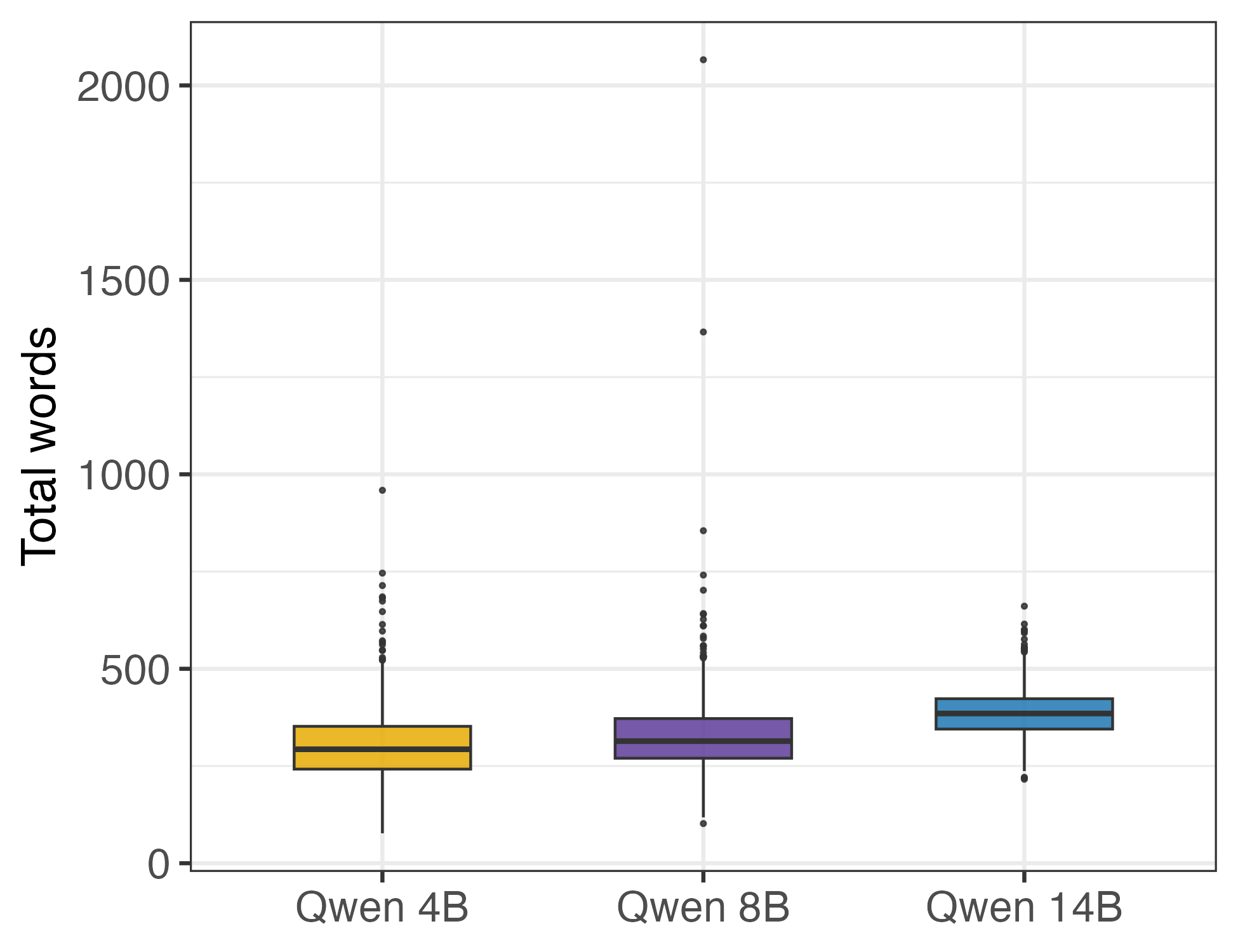}
     \caption{Distribution of total résumé length (word count) across LLM families (left) and Qwen 3 model sizes (right).}
     \label{fig:total_words_comparisons}
\end{figure}

\subsection{Screening Results}
\label{res:screening}

In the screening audit, we assess whether demographic information remains recoverable from generated résumés after anonymization and linguistic preprocessing.
Tables~\ref{tab:tfidf_gender_top_terms} and \ref{tab:tfidf_ethnicity_top_terms} report the most consistently selected TF--IDF terms across 30 repeated grouped LASSO runs, including selection frequency ($n$/30), mean number of selected features, average coefficient magnitude, and mean classifier performance (ACC, AUC).
We focus on the most selected terms only, as these reflect the strongest and most consistent leakage indicators.

\begin{table}[t]
\centering
\caption{
Most frequently selected TF--IDF lasso terms for gender prediction across 30 repeated grouped LASSO runs per model. Per term, we report the corresponding number of selected runs and mean coefficient across runs. Per model, we report mean selected features per run and mean classifier performance (Accuracy and AUC, with standard deviations (sd) over all runs).
}
\label{tab:tfidf_gender_top_terms}
\setlength{\tabcolsep}{3.5pt}
\begin{tabularx}{\linewidth}{Xcr@{\hspace{1em}}r@{\hspace{1em}}rr}
\toprule
Term & $n$ (/30)
& \multicolumn{1}{c}{Mean}
& \multicolumn{1}{c}{Mean selected}
& \multicolumn{2}{c}{Performance} \\
& &
\multicolumn{1}{c}{coefficient}
& \multicolumn{1}{c}{features}
& Acc
& \begin{tabular}[t]{@{}c@{}}AUC\\[-0.2em]{\scriptsize (sd)}\end{tabular} \\
\midrule
\texttt{ChatGPT} &  &  & 4.9 & 0.825 & \meansd{0.900}{0.046} \\
\cmidrule(lr){1-6}
fachkraft {\scriptsize specialist n.} & 30 & -0.991 (f.) &  &  &  \\
fach-mann/-frau {\scriptsize specialist m./f.} & 30 & 0.600 (m.) &  &  &  \\
\midrule
\texttt{Gemini} &  &  & 6.9 & 0.877 & \meansd{0.915}{0.034} \\
\cmidrule(lr){1-6}
fachkraft {\scriptsize specialist n.} & 30 & -1.541 (f.) &  &  &  \\
profi {\scriptsize professional n.} & 30 & 0.133 (m.) &  &  &  \\
\midrule
\texttt{Qwen 8B} {\scriptsize \texttt{top-k = 20}}
&  &  & 11.0 & 0.670 & \meansd{0.754}{0.038} \\
\cmidrule(lr){1-6}
fachkraft {\scriptsize specialist n.} & 30 & -0.554 (f.) &  &  &  \\
sport {\scriptsize sports} & 30 & 0.178 (m.) &  &  &  \\
ergebnisorientiert {\scriptsize goal oriented} & 30 & 0.130 (m.) &  &  &  \\
\bottomrule
\end{tabularx}

\smallskip
{\scriptsize m./f. = Positive coefficients indicate male (m.) association and negative female (f.) association.\par}
\end{table}

\paragraph{Gender Leakage.}

 Across all model families, gender information remains strongly recoverable despite anonymization and gender-neutralization (Table~\ref{tab:tfidf_gender_top_terms}).
 ChatGPT and Gemini exhibit particularly high leakage, achieving mean accuracies of 0.825 and 0.877 and AUC values of 0.900 and 0.915, respectively.
 Leakage for Qwen 8B is weaker but remains clearly above random guessing (ACC = 0.670, AUC = 0.754).
A notable result is the consistency of the selected linguistic signals.
Across all three model families, the term \emph{Fachkraft} (\emph{specialist}, neutral/non-gendered version n.) is selected in all 30 repeated runs and is consistently associated with female-associated résumés.
In contrast, male-associated résumés are characterized by alternative lexical choices.
For ChatGPT, the normalized term \emph{Fach-mann/-frau} (also translates to \emph{specialist}, but a gendered version m./f.) emerges in all 30 runs and exhibits a positive coefficient toward male-associated résumés, while Gemini instead selects \emph{Profi} (short for \emph{professional}, akin to \emph{pro}). 
Importantly, the identified terms are either naturally gender-neutral or have already undergone gender-neutralization during preprocessing.
The observed leakage therefore does not primarily arise from overt grammatical gender markers or explicitly gendered wording.
Rather, it reflects systematic differences in lexical choice, including the use of semantically overlapping alternatives.

At the same time, the Qwen 8B model exhibits some evidence of stereotypical lexical associations. Alongside the consistently selected gender-unspecific term \emph{Fachkraft}, Qwen repeatedly selects \emph{Sport} and \emph{ergebnisorientiert} (\emph{goal oriented}), the latter aligning with forms of agentic language commonly associated with masculine-coded communication in prior literature \cite{gaucher2011evidence}.
However, when extending the analysis beyond the LASSO-selected terms and comparing all generated résumés against established agentic and communal word lists, we do not observe consistent gendered language patterns across model families (see Appendix E).
This suggests that the observed demographic leakage is not driven by broad differences in agentic versus communal language use, but instead emerges through a comparatively small and stable subset of lexical preferences.

\begin{table}[t]
\centering
\caption{
Most frequently selected TF--IDF lasso terms for ethnicity prediction across 30 repeated grouped LASSO runs per model. Per term, we report the corresponding number of selected runs and mean coefficient across runs. Per model, we report mean selected features per run and mean classifier performance (Accuracy and AUC, with standard deviations (sd) over all runs).
}
\label{tab:tfidf_ethnicity_top_terms}
\setlength{\tabcolsep}{3.5pt}
\begin{tabularx}{\linewidth}{Xcr@{\hspace{1em}}r@{\hspace{1em}}rr}
\toprule
Term & $n$ (/30)
& \multicolumn{1}{c}{Mean}
& \multicolumn{1}{c}{Mean selected}
& \multicolumn{2}{c}{Performance} \\
& &
\multicolumn{1}{c}{coefficient}
& \multicolumn{1}{c}{features}
& Acc
& \begin{tabular}[t]{@{}c@{}}AUC\\[-0.2em]{\scriptsize (sd)}\end{tabular} \\
\midrule
\texttt{ChatGPT} &  &  & 1.7 & 0.500 & \meansd{0.499}{0.005} \\
\cmidrule(lr){1-6}
muttersprache {\scriptsize mother tongue} & 16 & -0.001 (t.) &  &  &  \\
\midrule
\texttt{Gemini} &  &  & 3.1 & 0.521 & \meansd{0.540}{0.027} \\
\cmidrule(lr){1-6}
austausch {\scriptsize exchange} & 24 & -0.023 (t.) &  &  &  \\
\midrule
\texttt{Qwen 8B} {\scriptsize \texttt{top-k = 20}}
&  &  & 1.6 & 0.500 & \meansd{0.503}{0.012} \\
\cmidrule(lr){1-6}
kulturell {\scriptsize cultural} & 6 & -0.001 (t.) &  &  &  \\
\bottomrule
\end{tabularx}

\smallskip
{\scriptsize g./t. = Positive coefficients indicate German (g.) association and negative Turkish (t.) association.\par}
\end{table}

\paragraph{Ethnicity Leakage.}

In contrast to gender, ethnicity-related leakage remains weak across all examined models (Table~\ref{tab:tfidf_ethnicity_top_terms}).
Classifier performance remains close to random guessing, with ChatGPT and Qwen 8B exhibiting essentially chance-level accuracy and AUC, while Gemini shows only a small deviation above random prediction (ACC = 0.521, AUC = 0.540).
Consistent with these results, selected ethnicity-related terms are substantially less stable than those observed for gender and exhibit negligible coefficient magnitudes.
ChatGPT most frequently selects \emph{Muttersprache} (\emph{mother tongue}), appearing in 16 of 30 runs, primarily in expressions such as \emph{Türkisch (Muttersprache)}.
Gemini most frequently selects \emph{Austausch} (\emph{exchange}), occurring in 24 runs and typically appearing in phrases such as \emph{interkultureller/internationaler Austausch} (\emph{intercultural/international exchange}). Qwen 8B most frequently selects \emph{kulturell} (\emph{cultural}), although only in 6 of 30 runs.
Nevertheless, despite little evidence of systematic linguistic separability between German- and Turkish-associated résumés, the recurring context of the selected terms indicates that they encode plausible ethnicity-related markers rather than entirely random variation.

\subsection{Scaling Across Open-Weight Models}
\label{res:qwen_scaling}

\begin{table}[h]
\centering
\caption{Leakage classifier performance across Qwen 3 model sizes and top-k decoding settings. Performance is averaged across 30 repeated runs.}
\label{tab:qwen_scaling_seed_performance}
\setlength{\tabcolsep}{8pt}
\begin{tabular}{llrrrrrr}
\toprule
Target & Setting & \multicolumn{2}{c}{4B} & \multicolumn{2}{c}{8B} & \multicolumn{2}{c}{14B} \\
\cmidrule(lr){3-4}\cmidrule(lr){5-6}\cmidrule(lr){7-8}
 & & Acc & AUC & Acc & AUC & Acc & AUC \\
\midrule
gender & \texttt{top-k=20} & 0.636 & 0.683 & 0.670 & 0.754 & 0.687 & 0.686 \\
 & \texttt{top-k=64} & 0.616 & 0.652 & 0.654 & 0.704 & 0.670 & 0.679 \\
\midrule
ethnicity & \texttt{top-k=20} & 0.500 & 0.499 & 0.500 & 0.503 & 0.566 & 0.579 \\
 & \texttt{top-k=64} & 0.519 & 0.529 & 0.500 & 0.500 & 0.542 & 0.557 \\
\bottomrule
\end{tabular}
\end{table}

Table~\ref{tab:qwen_scaling_seed_performance} summarizes leakage classifier performance across Qwen 3 model sizes and \texttt{top-k} decoding settings. Overall, gender leakage remains consistently more detectable than ethnicity leakage, which stays largely near chance level.
For gender, increasing \texttt{top-k} generally reduces both accuracy and AUC. Accuracy rises monotonically with model size, whereas AUC follows a reverse U-shape, peaking at the 8B model before declining slightly at 14B, though differences remain comparatively small.
Ethnicity prediction shows no consistent scaling pattern. While performance remains close to random guessing overall, the 14B model diverges most strongly, reaching an accuracy of 0.566 and an AUC of 0.579 at \texttt{top-k=20}.

\section{Discussion \& Conclusion}
\label{discussion}
Our results show that German-language LLM-generated résumés can encode highly recoverable gender information despite explicit anonymization and gender-neutralization procedures.
Downstream TF--IDF leakage classifiers achieve high gender prediction performance for ChatGPT and Gemini and moderate performance for Qwen 3, indicating that demographic information remains recoverable even after removing explicit demographic indicators.
Importantly, the observed leakage is not primarily driven by overtly gendered language, but rather by subtle differences in the usage of semantically equivalent, formally gender-neutral terms.
Most notably, the formally gender-neutral term \emph{Fachkraft} (\emph{specialist n.}), which can refer to both men and women in German and is commonly used as an inclusive alternative to gender-marked occupational nouns, is strongly associated with female-associated résumés across all model versions.
However, semantically overlapping but gender-marked versions of the same term in the form of \emph{Fachmann/-frau} (\emph{specialist m./f.}) occur predominantly in male-associated résumés.
In contrast, we find substantially weaker and largely inconsistent evidence of ethnicity-related leakage.
Linguistic differences between German- and Turkish-associated résumés remain small as ethnicity classifiers perform close to random guessing during the audit.

The scaling analysis across Qwen 3 model sizes supports the main findings.
Gender leakage remains detectable across all model sizes and top-k settings, whereas ethnicity prediction stays near chance level, mirroring the pattern observed for ChatGPT and Gemini.
Gender leakage does not increase monotonically with scale: while accuracy rises with model size, AUC peaks at 8B and declines at 14B.
A higher top-k generally reduces gender prediction performance, whereas no consistent pattern emerges for ethnicity.
Together, these findings suggest that demographic leakage reflects model-specific lexical preferences rather than scale alone.

These results have important implications for AI-assisted hiring pipelines.
Existing fairness interventions in résumé screening often rely on anonymization procedures intended to remove protected attributes from application documents \cite{saldivar2025synthetic,parasurama-sedoc-2022-gendered}.
Our findings therefore suggest that anonymization alone may not fully eliminate demographic information when lexical alternatives themselves encode socially patterned variation.
Moreover, even seemingly unrelated résumé characteristics, including formatting choices or references to external profiles such as LinkedIn, have been shown to influence automated screening outcomes \cite{rhea_resume_format}.
This highlights the need for heightened awareness of seemingly neutral differences, such as gender-neutral terms.
Importantly, the observed leakage patterns appear closely tied to properties of the German language, where semantically overlapping but differently gender-marked lexical alternatives coexist \cite{motschenbacher2014grammatical}.
To our knowledge, similar effects remain comparatively understudied in related résumé-screening research, which has focused primarily on English-language settings.
Accordingly, findings from this literature may not transfer directly to multilingual contexts, underscoring the need for language-specific fairness evaluations in AI-assisted recruitment systems.

Beyond theoretical implications, these findings carry practical implications for both applicants and hiring platforms. For applicants who utilize LLMs in creating and refining application materials, omitting names and other direct demographic markers from prompts may reduce the risk of unintended demographic encoding in the output. For companies and hiring platforms, our results suggest that simple anonymization of final documents is insufficient to fully remove demographic information. A preliminary step that detects and neutralizes gender-associated lexical variation in applicant materials or summarized texts may reduce the risk of demographically biased screening outcomes.

Our findings should be interpreted in light of several design choices and limitations.
First, the results depend on specific experimental decisions, including the selected names, prompts, model families, and decoding settings.
Further, the present audit focuses on controlled LLM-generated résumé text rather than naturally occurring application materials.
While our design enables systematic variation of demographic markers and controlled comparison across models, it remains unclear to what extent similar demographic leakage persists, is weakened, or may even become amplified in real-world résumé writing practices, where applicants may combine human authorship with varying degrees of AI assistance.
Finally, while we demonstrate strong demographic recoverability, recoverable demographic information should not be interpreted as direct evidence of discriminatory hiring outcomes.
Rather, demographic leakage constitutes a prerequisite for potential downstream bias.
Whether and how these linguistic signals ultimately influence real hiring decisions depends on the specific screening systems, feature representations, and decision processes used by employers and may therefore vary substantially across contexts.

\section*{Acknowledgements}

We thank the German job-matching company Chemistree GmbH for their cooperation in this work and for taking on the role of project partners as part of a student consulting project. 

\section*{Ethical Statement}
\label{sec:ethical_statement}
The matching profiles were provided to the authors in anonymized form by Chemistree GmbH, which is contracted as an IT service provider by messerocks GmbH, the organization operating the matching service. Users had explicitly consented to the use of their questionnaire responses for matching purposes and to their transfer to Chemistree GmbH as an IT service provider. The corresponding data processing agreement required processing in accordance with the EU General Data Protection Regulation and permitted Chemistree GmbH to use anonymized data for development and research purposes. The study used only anonymized structured questionnaire profiles without direct personal identifiers, gender or ethnicity labels, employers, institutions, or dates. Demographic variation was introduced exclusively through externally sampled synthetic names.

\bibliographystyle{splncs04}
\bibliography{bibliography}
\newpage

\appendix
\hypersetup{pageanchor=false}
\setcounter{page}{1}
\setcounter{section}{0}
\providecommand{\theHpage}{}
\renewcommand{\theHpage}{appendix.\arabic{page}}
\renewcommand{\theHsection}{appendix.\arabic{section}}

\section{Appendix}
\titlerunning{Appendix: Fairness Beyond Anonymization}
\subsection{Matching Profile Data}
The anonymized data comprises 23 items in single-choice, multiple-choice, and hierarchical multiple-choice formats. An overview of all questionnaire items is provided in Table~\ref{tab:chemistree_variables}.
One-hot encoding all responses yields $K= 1{,}058$ binary variables, distributed across categories as follows: preferences account for 50.8\%, hard skills for 40.6\%, and soft skills for 8.6\% of all variables.
Across fully completed profiles, applicants select on average 100.19 responses (9.5\% of all options). 
The subsample used for the experiments exhibits a slightly lower average of 95.27 (9.0\%).

\begin{scriptsize}
\renewcommand{\arraystretch}{1.4}
\begin{longtable}{@{} >{\RaggedRight}p{3cm} l @{\hspace{1em}} >{\RaggedRight}p{3.8cm} @{\hspace{1em}}>{\RaggedRight}p{3cm} @{}}
\caption{Overview of matching profile questionnaire items. Translated from German; illustrative selection of response options. Items are grouped into hard skills, soft skills, and preferences.}
\label{tab:chemistree_variables} \\
\toprule
\textbf{Item Name} & \textbf{Type} & \textbf{Answers} & \textbf{Subanswers} \\ \midrule
\endfirsthead

\multicolumn{4}{c}{
} \\
\toprule
\textbf{Item Name} & \textbf{Type} & \textbf{Answers} & \textbf{Subanswers} \\ \midrule
\endhead

\midrule \multicolumn{4}{r}{{Continued on next page}} \\
\endfoot

\bottomrule
\multicolumn{4}{p{\linewidth}}{%
\textit{\textbf{Note:} SINGLE (single choice); MULTI (multiple choice); MULTI\_H (hierarchical multiple choice).
For MULTI\_H, the Answer column provides the top-level answer. The Subanswer column contains either fixed options (e.g., language proficiency) or options specific to the top-level selection (e.g., sub-departments).}}
\endlastfoot

\multicolumn{4}{l}{\textbf{Hard Skills}} \\ \midrule
Industry Experience & MULTI & Open to all; Agency / Advertising / Marketing / PR; Plant and mechanical engineering / industry; ... & -- \\
Professional Profile & MULTI\_H & Agricultural / Forestry / Environmental Sciences; Construction / Architecture / Surveying; ... & Entry; Professional; Management; ... \\
Education Background & MULTI & Abitur; Vocational training; Bachelor's Degree; ... & -- \\
Language (German) & SINGLE & None; Proficiency between A1--C2 & -- \\
Language (English) & SINGLE & None; Proficiency between A1--C2 & -- \\
Further languages & MULTI\_H & Albanian; Arabic; ... & Proficiency between A1--C2 \\ \midrule

\multicolumn{4}{l}{\textbf{Soft Skills}} \\ \midrule
Task Description & MULTI & Abstracting; Weighing; Analyzing; ... & -- \\
Personal skills & MULTI & Adaptability; Appearance; Endurance; ... & -- \\
Social-comm. skills & MULTI & Empathy; Ability to motivate others; ... & -- \\
Activity-oriented skills & MULTI & Analysis capability; Judgment; Ability to delegate; ... & -- \\ \midrule

\multicolumn{4}{l}{\textbf{Preferences}} \\ \midrule
Desired industry & MULTI & Open to all; Agency / Advertising / Marketing / PR; ... & -- \\
Preferred company size & MULTI & From Micro ($< 9$) to Major ($> 1000$) Enterprises & -- \\
Topics/Departments & MULTI\_H & Banking; Design; ... & E.g. for Banking: Credit; Digital Finance; ... \\
Internal department & MULTI & Admin; Agile; Analysis; ... & -- \\
Business environment & MULTI & New-work; Start-up; Family; ... & -- \\
Place of Work & MULTI\_H & Within GER: Bundesländer; Outside of GER: EU/USA/Other & Within GER: Specific cities; Otherwise: Countries \\
Remote activity & SINGLE & Yes; No & -- \\
Scope of the job & MULTI & Part time; Full; Freelance; ... & -- \\
Travel activity & SINGLE & High; Limited; No & -- \\
Team Role Preferences & MULTI & Advocatus diaboli; Integrator; Coordinator; ... & -- \\
Important Team Values & MULTI & Open-mindedness; Enthusiasm; Effectiveness; ... & -- \\
Company Score Points & MULTI & Equal Pay; Flat Hierarchies; New Work; ... & -- \\
Gross annual salary & MULTI & No statement; Unpaid; categories between $<$6.24k and $>$150k EUR & -- \\
\end{longtable}
\end{scriptsize}

\subsection{Full Name List}

\begin{table}[ht]
\centering
\caption{Complete list of sampled names grouped by ethnicity and gender.}
\label{tab:full_name_list}

\scriptsize
\setlength{\tabcolsep}{13pt}
\renewcommand{\arraystretch}{1.05}

\begin{tabular}{llll}
\toprule
\multicolumn{2}{c}{\textbf{German}} & \multicolumn{2}{c}{\textbf{Turkish}} \\
\cmidrule(lr){1-2} \cmidrule(lr){3-4}
\textbf{Male} & \textbf{Female} & \textbf{Male} & \textbf{Female} \\
\midrule
Marcel Popp       & Ramona Popp       & Emre Eren       & Havva Eren \\
Philipp Stoll     & Mareen Stoll      & Onur Ayyildiz   & Kübra Ayyildiz \\
Kai Lindner       & Carina Lindner    & Orhan Turan     & Cansu Turan \\
Fabian Kern       & Britta Kern       & Fikret Akman    & Necla Akman \\
Marco Neubauer    & Jennifer Neubauer & Yunus Sönmez    & Gül Sönmez \\
Rene Schmitt      & Ute Schmitt       & Şahin Acar      & Suna Acar \\
Daniel Kuhn       & Daniela Kuhn      & Levent Karakaya & Dilek Karakaya \\
Michael Ackermann & Sonja Ackermann   & Suat Mutlu      & Özlem Mutlu \\
Jörg Engel        & Nicole Engel      & Ahmet Akay      & Hilal Akay \\
Henrik Kohl       & Angelika Kohl     & Adem Şeker      & Hatice Şeker \\
\bottomrule
\end{tabular}
\end{table}

\subsection{Full Prompt}
The used prompt was designed to support a consistent output format and encourage the synthesis of specific details that are commonly included in résumés.

\begin{paragraph}{\textbf{Original German Prompt.}}
\begin{verbatim}
[SYSTEM]
Du bist ein professioneller Karriere-Coach.  
Erstelle einen professionellen Lebenslauf (max. 1 Seite)
in deutscher Sprache auf Basis von Fragebogen-Daten. 
Keine Platzhalter (z. B. [NAME]); verwende realistische,
konsistente Angaben. 
Wenn Informationen fehlen, ergänze realistisch.
Halte Dich an übliche Konventionen.

Ausgabeformat:  
- Gib ausschließlich ein einzelnes JSON-Objekt aus.
  Kein erläuternder Text, keine Einleitung, keine Codeblöcke.
- Verwende genau die folgenden Top-Level-Schlüssel in
  dieser Reihenfolge:  
  - 01_persoenliche_daten  
  - 02_profil  
  - 03_faehigkeiten  
  - 04_berufserfahrung  
  - 05_ausbildung  
  - 06_skills  
  - 07_sprachen  
  - 08_interessen 
  - 09_angestrebte_position 
  - 10_cover_letter_snippet

[USER] 
Hier sind die Fragebogen-Daten (JSON):
```json
{profile_qa_json}
```

Aufgabe:
Erstelle einen Lebenslauf-JSON für: {first_name} {last_name},
{age} Jahre auf Basis der Antworten des Fragebogens.

Beispiel-Ausgabeformat:
{{"01_persoenliche_daten":..., "02_profil":..., ...}}
\end{verbatim}
\end{paragraph}

\newpage

\paragraph{\textbf{Translated English Prompt.}}
\begin{verbatim}
[SYSTEM]
You are a professional career coach.
Create a professional resume (max. 1 page)
in German based on questionnaire data.
No placeholders (e.g. [NAME]); use realistic,
consistent details.
If information is missing, fill it in realistically.
Follow standard resume conventions.

Output format:
- Output only a single JSON object. No explanatory text,
  no introduction, no code blocks.
- Use exactly the following top-level keys
  in this order:
  - 01_personal_data 
  - 02_profile
  - 03_competences  
  - 04_work_experience  
  - 05_education  
  - 06_skills 
  - 07_languages
  - 08_interests 
  - 09_desired_position
  - 10_cover_letter_snippet
  
[USER]
Here are the questionnaire data (JSON):
```json
{profile_qa_json}
```

Task:
Create a resume-JSON for: {first_name} {last_name},
age {age} on the basis of the questionnaire answers.

Example output format: 
{"01_personal_data":..., "02_profile":..., ...}
\end{verbatim}

\subsection{Compute Infrastructure}
Qwen 3 models were served locally using vLLM on a single NVIDIA RTX A6000 GPU (48\,GB VRAM) in bfloat16 precision. GPT-4o-mini and Gemini 2.5 Flash-Lite were accessed via their respective APIs. The project was implemented in Python (version 3.12.2) and R (version 4.5.1).

\newpage

\subsection{Résumé Results}

\begin{figure}[htbp]
\centering
\includegraphics[width=0.85\textwidth]{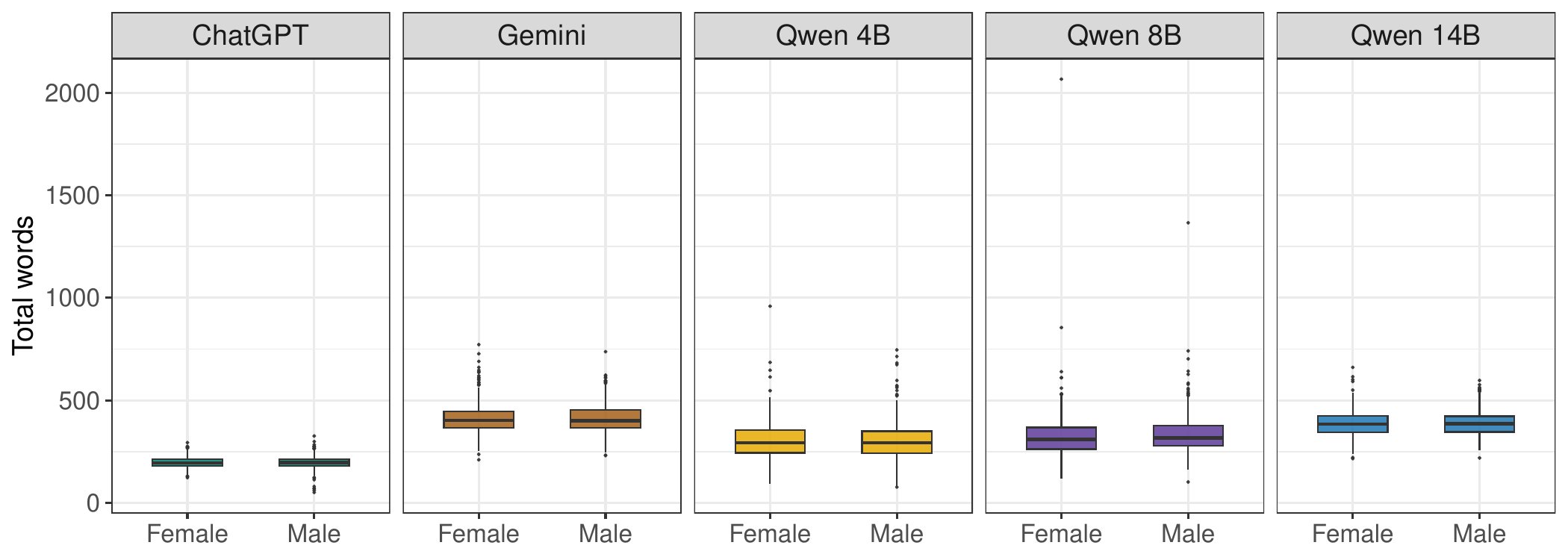}
\caption{Distribution of total résumé length (word count) by gender across all model families. No systematic differences in résumé length are observed between male- and female-associated résumés.}
\label{fig:words_gender}
\end{figure}

\begin{figure}[htbp]
\centering
\includegraphics[width=0.85\textwidth]{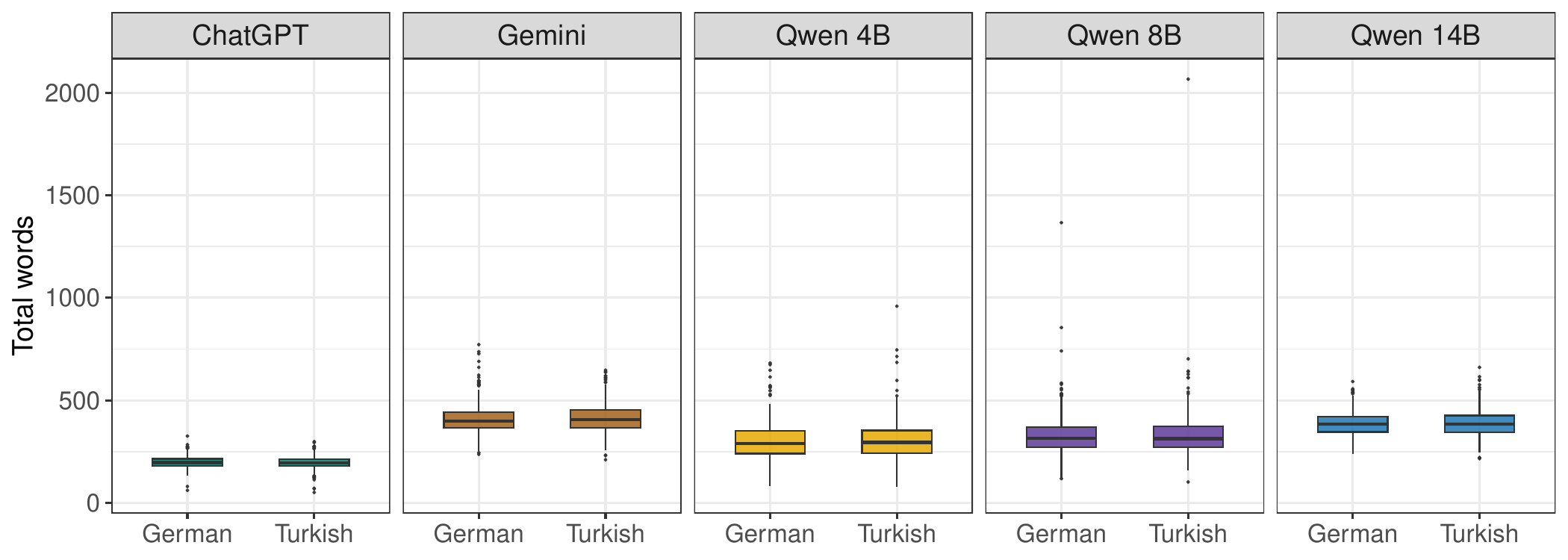}
\caption{Distribution of total résumé length (word count) by ethnicity across all model families. Résumé length remains broadly comparable between German- and Turkish-associated résumés, indicating no substantial ethnicity-related differences in output length.}
\label{fig:words_ethnicity}
\end{figure}

\newpage 

\subsubsection{Agentic \& Communal Language Across Gender}

\begin{figure}[htbp]
\centering
\includegraphics[width=0.85\textwidth]{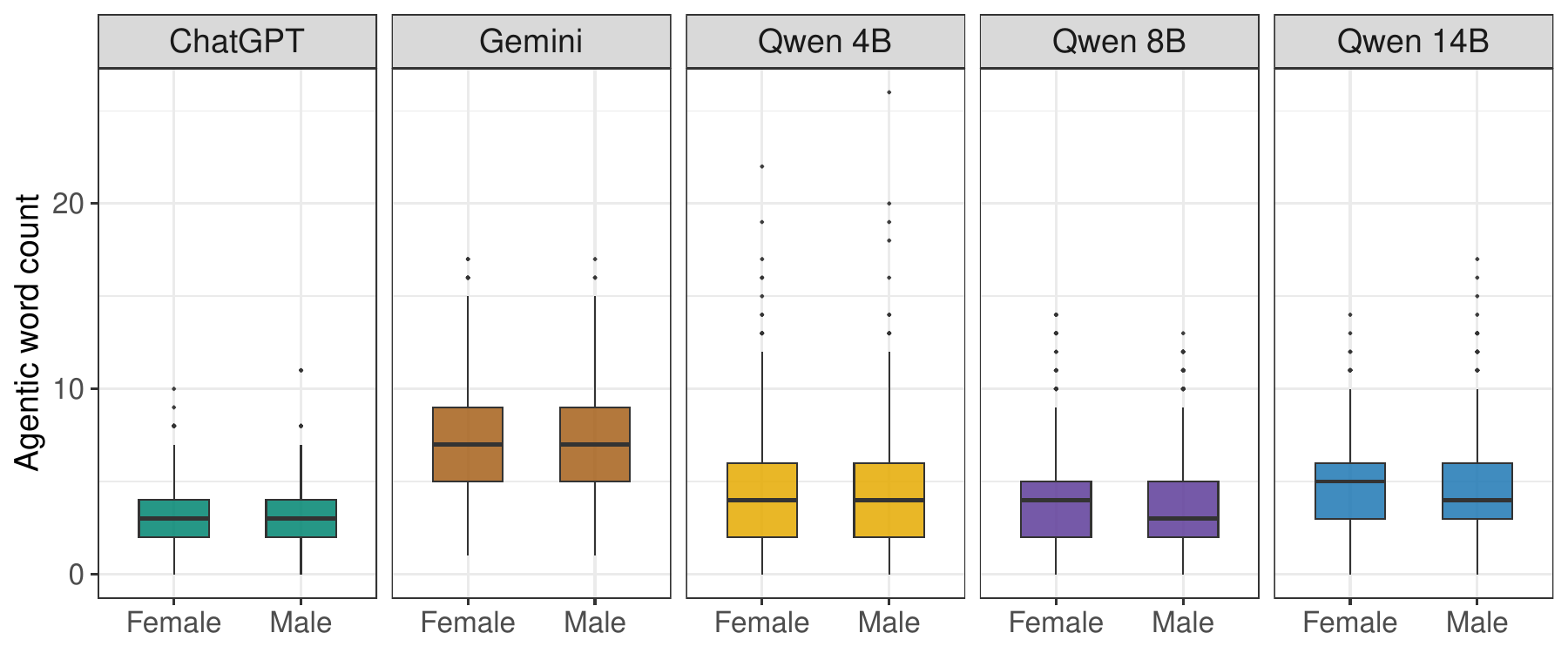}
\caption{Distribution of agentic word counts by gender across all model families.}
\label{fig:agentic_gender}
\end{figure}

\begin{figure}[htbp]
\centering
\includegraphics[width=0.85\textwidth]{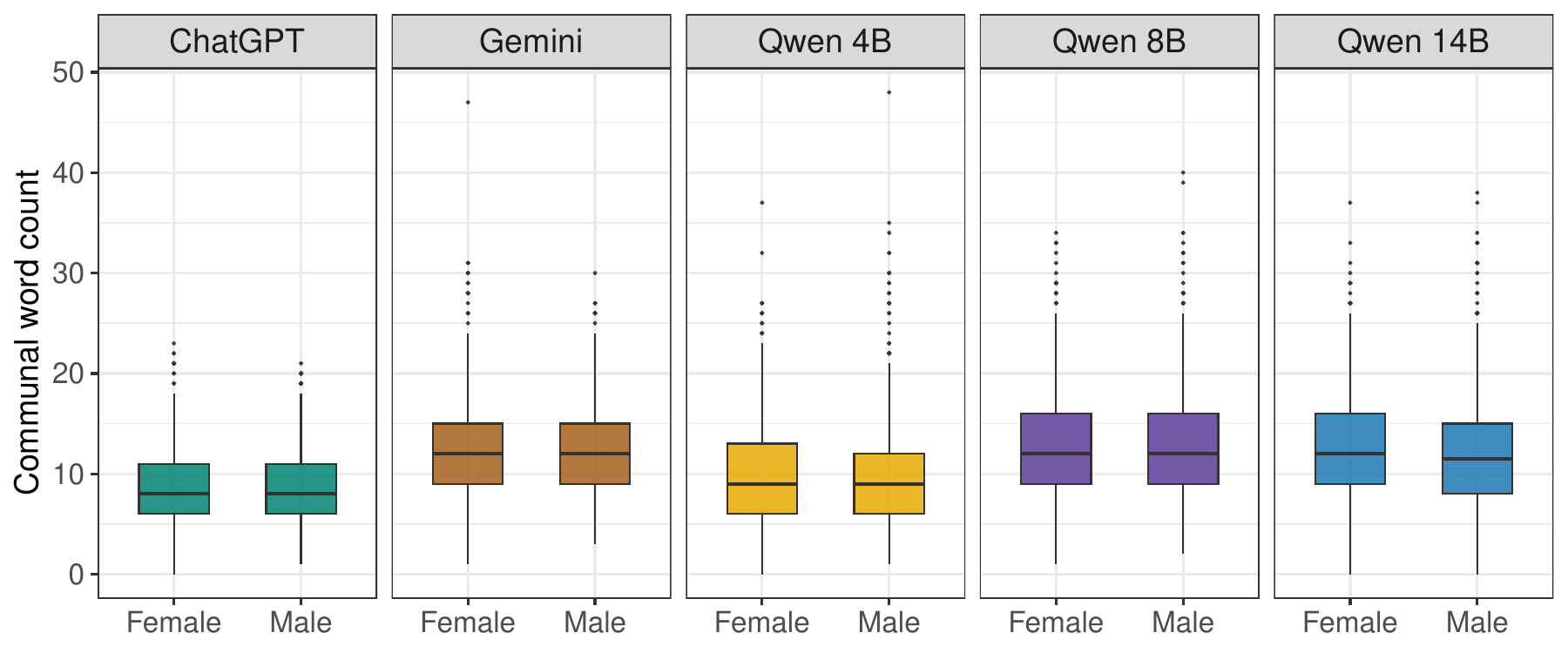}
\caption{Distribution of communal word counts by gender across all model families.}
\label{fig:communal_gender}
\end{figure}

Literature on gendered language in recruitment materials repeatedly observes differences in the use of agentic words (such as assertiveness, ambition, and competitiveness), associated with masculine traits, and communal words (such as cooperation, warmth, and social orientation), associated with feminine traits. As shown in Figures~\ref{fig:agentic_gender} and~\ref{fig:communal_gender}, we do not find such differences in the LLM-generated résumés on a broader level. German agentic and communal word lists were obtained from the Technical University of Munich.\footnote{German agentic and communal word lists: \url{https://www.msl.mgt.tum.de/rm/third-party-funded-projects/projekt-fuehrmint/gender-decoder/wortlisten/}}

\end{document}